\documentclass[conference]{IEEEtran}
\IEEEoverridecommandlockouts
\usepackage{cite}
\usepackage{amsmath,amssymb,amsfonts}
\usepackage{algorithmic}
\usepackage{graphicx}
\usepackage{textcomp}
\usepackage{xcolor}
\usepackage{colortbl}
\usepackage{tabularx}
\usepackage{relsize}
\usepackage{pifont}
\usepackage{booktabs} 
\usepackage{multirow}
\usepackage{multicol}
\usepackage{adjustbox}
\usepackage{float}
            
\usepackage{textcomp}
\renewcommand{\baselinestretch}{1.0} 

\usepackage{amsmath,amsfonts,amssymb}
\usepackage{graphicx}
\usepackage[colorlinks=true, allcolors=blue]{hyperref}

\def\BibTeX{{\rm B\kern-.05em{\sc i\kern-.025em b}\kern-.08em
    T\kern-.1667em\lower.7ex\hbox{E}\kern-.125emX}}
\begin{document}

\title{All-in-SAM: from Weak Annotation to Pixel-wise Nuclei Segmentation with Prompt-based Finetuning\\
\thanks{This research was supported by NIH R01DK135597 (Huo), NSF CAREER 1452485, NSF 2040462,NCRR Grant UL1-01 (now at NCATS Grant 2 UL1 TR000445-06), NVIDIA hardware grant, resources of ACCRE at Vanderbilt University}
}

\author{\IEEEauthorblockN{Can Cui}
\IEEEauthorblockA{\textit{Computer Science} \\
\textit{Vanderbilt University}\\
Nashville, USA \\
can.cui.1@vanderbilt.edu \\ 
}
\and
\IEEEauthorblockN{Ruining Deng}
\IEEEauthorblockA{\textit{Computer Science} \\
\textit{Vanderbilt University}\\
Nashville, USA \\
r.deng@vanderbilt.edu \\ 
}
\and
\IEEEauthorblockN{Quan Liu}
\IEEEauthorblockA{\textit{Computer Science} \\
\textit{Vanderbilt University}\\
Nashville, USA \\
quan.liu@vanderbilt.edu \\ 
}
\and
\IEEEauthorblockN{Tianyuan Yao}
\IEEEauthorblockA{\textit{Computer Science} \\
\textit{Vanderbilt University}\\
Nashville, USA \\
tianyuan.yao@vanderbilt.edu \\
}
\and
\IEEEauthorblockN{Shunxing Bao}
\IEEEauthorblockA{\textit{Electrical and Computer Engineering} \\
\textit{Vanderbilt University}\\
Nashville, USA \\
shunxing.bao@vanderbilt.edu \\
}
\and
\IEEEauthorblockN{Lucas W. Remedios}
\IEEEauthorblockA{\textit{Computer Science} \\
\textit{Vanderbilt University}\\
Nashville, USA \\
lucas.w.remedios@vanderbilt.edu \\
}
\and
\IEEEauthorblockN{Bennett A. Landman}
\IEEEauthorblockA{\textit{Electrical and Computer Engineering} \\
\textit{Vanderbilt University}\\
Nashville, USA \\
bennett.landman@vanderbilt.edu \\
}
\and
\IEEEauthorblockN{Yucheng Tang}
\IEEEauthorblockA{\textit{NVIDIA Cooperation} \\
Redmond, WA, USA \\
yucheng.tang@vanderbilt.edu \\
}
\and
\IEEEauthorblockN{Yuankai Huo}
\IEEEauthorblockA{\textit{Computer Science} \\
\textit{Vanderbilt University}\\
Nashville, USA \\
yuankai.huo@vanderbilt.edu}
}

\maketitle

\begin{abstract}
The Segment Anything Model (SAM) is a recently proposed prompt-based segmentation model in a generic zero-shot segmentation approach. With the zero-shot segmentation capacity, SAM achieved impressive flexibility and precision on various segmentation tasks. However, the current pipeline requires manual prompts during the inference stage, which is still resource intensive for biomedical image segmentation. In this paper, instead of using prompts during the inference stage, we introduce a pipeline that utilizes the SAM, called all-in-SAM, through the entire AI development workflow (from annotation generation to model finetuning) without requiring manual prompts during the inference stage. Specifically, SAM is first employed to generate pixel-level annotations from weak prompts (e.g., points, bounding box). Then, the pixel-level annotations are used to finetune the SAM segmentation model rather than training from scratch. Our experimental results reveal two key findings: 1) the proposed pipeline surpasses the state-of-the-art (SOTA) methods in a nuclei segmentation task on the public Monuseg dataset, and 2) the utilization of weak and few annotations for SAM finetuning achieves competitive performance compared to using strong pixel-wise annotated data. 

\end{abstract}


\section{Introduction}
The foundation models have recently been proposed as a powerful segmentation model~\cite{brown2020language,openai2023gpt4}. Segment Anything Model (SAM), as an example, was trained by millions of images to achieve a generic segmentation capability~\cite{kirillov2023segment}. SAM can automatically segment a new image, and it also accepts the prompts input of foreground/background points or the box regions for better segmentation~\cite{deng2023segment,ma2023segment,wu2023medical,zhang2023input}. However, recent studies have revealed SAM's limited performance in specific domain tasks, such as medical image segmentation, particularly when an insufficient number of prompts are available~\cite{deng2023segment}. The main reason is that medical data was rare to see in the training set of SAM while the medical segmentation tasks always in requirement of higher professional knowledge than natural image segmentation~\cite{huo2021ai}. 

Finetuning strategy utilizes the power of the generic model in detecting low-level and general image patterns but adjusts the final segmentation based on the characteristics and high-level understanding of downstream tasks, which provides a promising solution in adapting the generic segmentation model to downstream tasks. Previous approaches~\cite{wu2023medical, chen2023sam} have proposed finetuning methods to improve SAM's performance in downstream tasks. However, these methods mostly require complete data annotation for finetuning and did not explore the impact of weak annotation and few training data on the finetuning of the pretrained SAM model. 

Nuclei segmentation is a crucial task in biomedical research and clinical applications, but manual annotation of nuclei in whole slide images (WSIs) is time-consuming and labor-intensive. Previous works attempted to automatically segment nuclei with supervised learning~\cite{kong2020nuclear, hu2019mc}. More recently, some methods used self-supervised learning to further improve the model performance~\cite{xie2020instance, sahasrabudhe2020self}. SAM has great potential to benefit nuclei segmentation if it can be adapted appropriately. This paper investigates the performance of transferring the SAM to nuclei segmentation. A previous study has indicated that SAM performed poorly in nuclei segmentation without box/point information as manual prompts, but achieved promising segmentation when the bounding box of every nuclei was provided as the prompt in the inference stage ~\cite{deng2023segment}. However, manually annotating all the boxes during inference remains time-consuming. To address this issue, we introduce a pipeline for label-efficient finetuning of SAM, with no requirement for annotation prompts during inference. Also, instead of relying on complete annotations for finetuning, we leverage weak annotations to reduce annotation costs further while achieving comparable segmentation performance to SOTA methods. 

\begin{figure}[t]
\begin{center}
\includegraphics[width=0.90\linewidth]{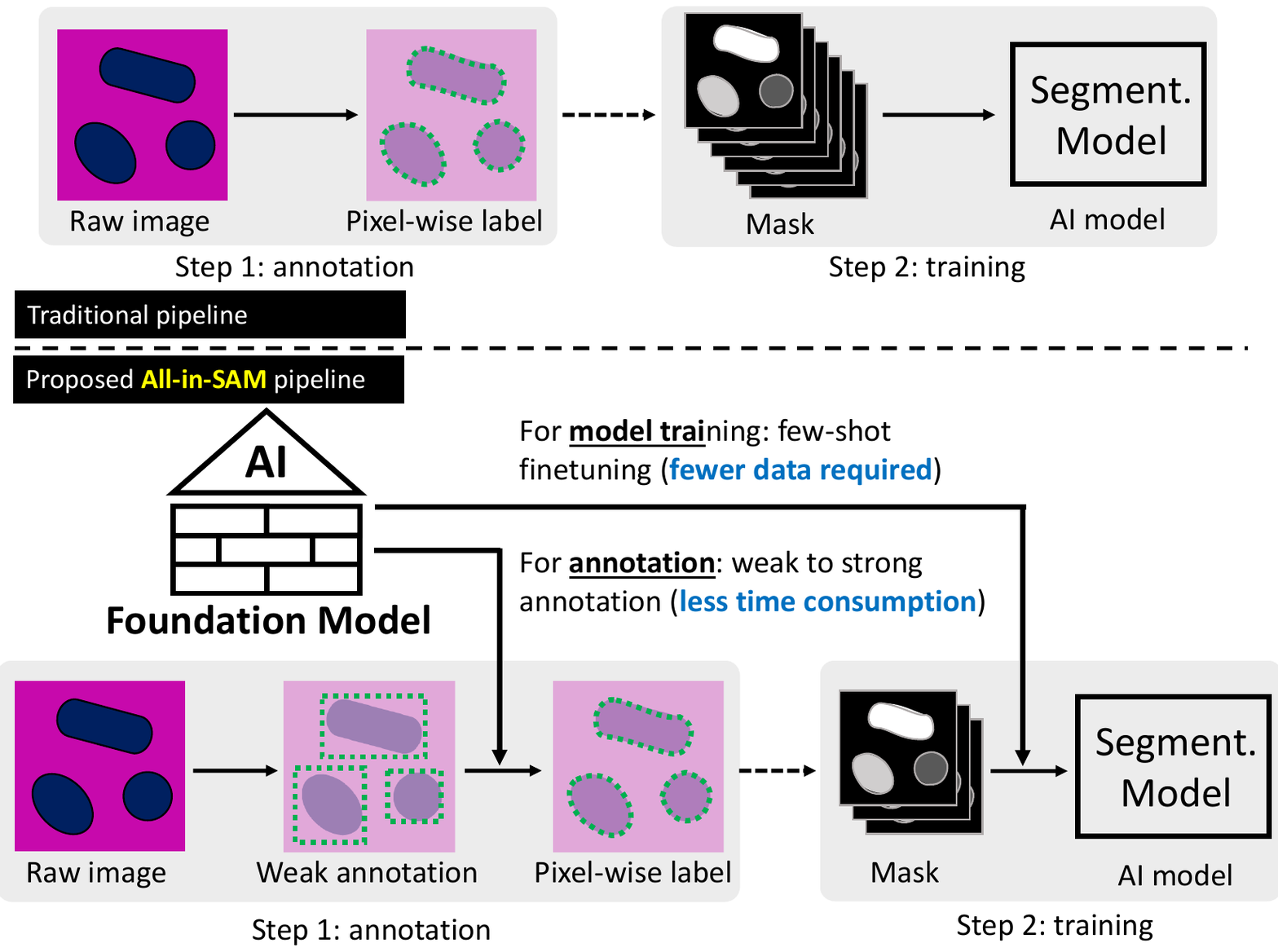}
\end{center}
   \caption{This figure shows the overall idea of the proposed All-in-SAM pipeline. First, the AI foundation model SAM is used in the annotation phase to convert weak annotations (bounding boxes) to strong annotations (pixel-wise labels), which reduces the time consumption during the labeling process. Then, the SAM model is fine-tuned with fewer strong annotations. The ultimate goal of the All-in-SAM pipeline is to enable efficient few-shot and weak annotation for AI model adaptations.}
\label{fig0:concept}
\end{figure}

In this work, we proposed the All-in-SAM pipeline, utilizing the pretrained SAM for annotation generation and model finetuning. Also, no manual prompts are required during the inference stage (Fig.~\ref{fig0:concept}). The contribution of this work can be summarized into two points:  


1) Utilization of weak annotations for cost reduction: Rather than relying exclusively on fully annotated data for finetuning, we demonstrate the effectiveness of leveraging weak annotations and the pretrained SAM. This approach helps to minimize annotation costs while achieving segmentation performance that is comparable to the current state-of-the-art methods.

2) Development of a pipeline for label-efficient finetuning: We propose a method that allows SAM to be finetuned for nuclei segmentation without the requirement of manual prompts during inference. This significantly reduces the time and effort involved in manual annotation.

Overall, this work aims to enhance the application of SAM in nuclei segmentation by addressing the annotation burden and cost issues through label-efficient finetuning and the utilization of weak annotations.

\section{Method} \label{Method}

\begin{figure*}[t]
\begin{center}
\includegraphics[width=0.90\linewidth]{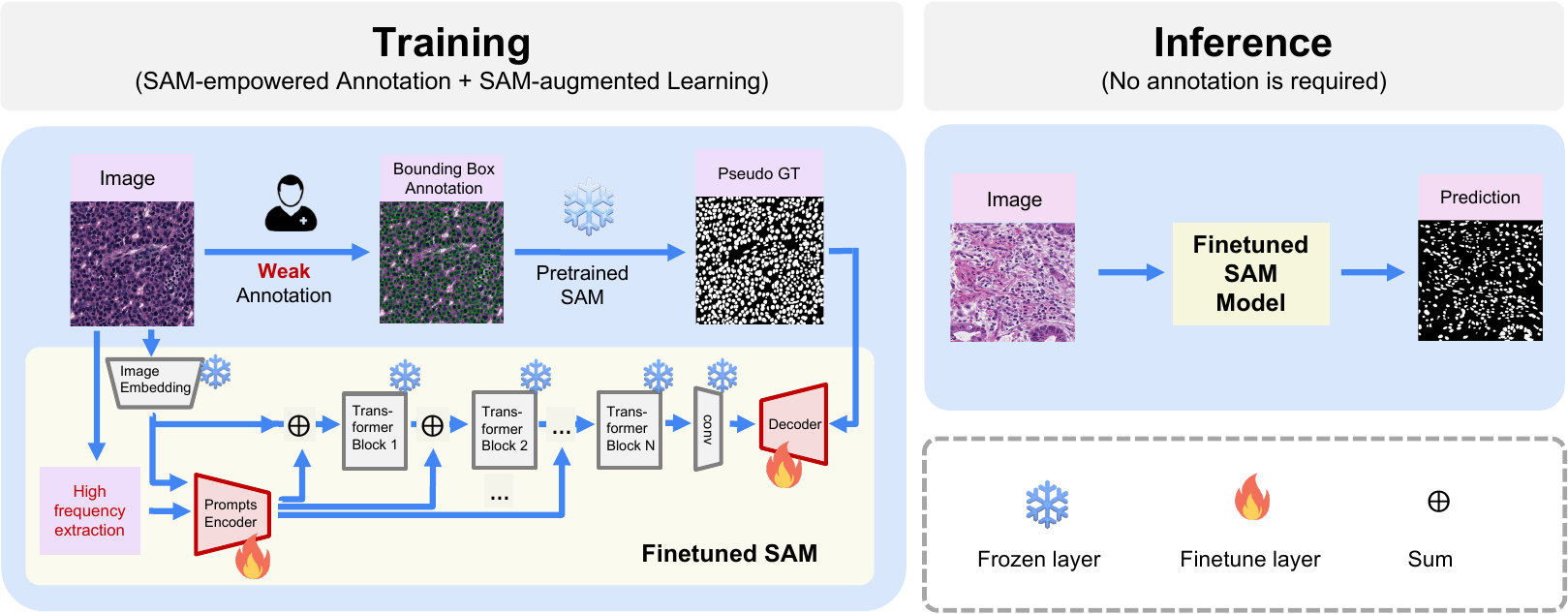}
\end{center}
   \caption{The proposed pipeline using weak and few annotations for nuclei segmentation. In the training stage, only the bounding boxes (in green color) of nuclei were provided as the weak annotation label to generate the approximate segmentation masks. Then, with the supervision of the approximate segmentation masks, the prompt-based finetuning was applied to the pretrained SAM model. In the inference stage, nuclei can be segmented directly from images without box prompts.}
\label{fig1:Pipeline}
\end{figure*}

\subsection{Overview} \label{Overview}
Motivated by the promising performance of the SAM model in interactive segmentation tasks with sparse prompts and the potential for finetuning, we propose a segmentation pipeline that leverages weak and limited annotations and apply this pipeline to the nuclei segmentation task.

The proposed pipeline consists of two main stages: SAM-empowered annotation and SAM finetuning. In the first stage, we utilize the pretrained SAM model to generate high-quality approximate nuclei masks for pathology images. This is achieved by providing the bounding boxes of nuclei as input to the pretrained SAM model. These approximate masks serve as initial segmentation masks for the nuclei. In the second stage, the generated approximate masks are employed to finetune the SAM model, which allows the model to adapt and refine its segmentation capabilities specifically for nuclei segmentation. The proposed pipeline is displayed in Fig.~\ref{fig1:Pipeline}. Two stages are introduced in detail in~\ref{SAM-empowered annotation} and~\ref{SAM-finetuning}.

Furthermore, we evaluate the performance of the model when only a small number of annotated data for downstream tasks. By decreasing the number of annotated samples, we aim to reduce annotation labor while still achieving satisfactory segmentation results.

\subsection{SAM-empowered annotation}  \label{SAM-empowered annotation}
The SAM model consists of three key components: the prompt encoder, the image encoder, and the mask decoder. The image encoder utilizes the Vision Transformer (ViT) as its backbone. The prompt encoder can take two forms: sparse or dense. In the sparse form, prompts can be in the form of points, boxes, or text, whereas in the dense form, prompts are represented as a grid or mask. The encoded prompts are then added to the image representation for the subsequent mask decoding process. In a previous study~\cite{deng2023segment}, it was observed that when only automatically generated dense prompts were used, nuclei segmentation sometimes failed to produce satisfactory results. However, significant improvement was achieved when weak annotations such as points or boxes were provided during the segmentation inference. Notably, when the bounding box of the nucleus was available as a weak annotation, the segmentation achieved a dice value of 0.883 in the public Monuseg dataset~\cite{kumar2017dataset}, significantly surpassing the results obtained from supervised learning methods. It indicates that SAM has strong capabilities in edge detection, enabling clear detection of nuclei boundaries within focus regions. This makes it a potential tool to generate precise approximate masks, which can enhance supervised learning approaches with lower annotation costs.

\subsection{SAM-finetuning} \label{SAM-finetuning}
SAM has been trained on a large dataset for generic segmentation tasks, giving it the ability to perform well in general segmentation. However, when applied to specific tasks, SAM may exhibit suboptimal performance or even fail. Nonetheless, if the knowledge accumulated by SAM can be transferred to these specific tasks, it holds great potential for achieving better performance compared to training the model from scratch using only downstream task data, especially when the available annotated data for the downstream task is limited. 

To optimize the transfer of knowledge, rather than finetuning the entire large pretrained model, a more effective and efficient approach is to selectively unfreeze only the last few layers. However, in our experiments, this approach still yielded inferior results compared to some baselines. Recently, there has been growing attention in the natural language processing community toward the use of adapters as an effective tool for finetuning models for different downstream tasks by leveraging task-specific knowledge~\cite{houlsby2019parameter}. In line with this, Chen et al.~\cite{chen2023sam} successfully adapted prompt adapters~\cite{liu2023explicit} in the finetuning process of SAM. Specifically, they automatically extracted and encoded the texture information of each image as handcrafted features, which were then added to multiple layers in the encoder. Additionally, the proposed prompt encoder, along with the unfrozen lightweight decoder, became learnable during the finetuning process. Following their work, we implement such finetuning strategy in the nucleus segmentation task, but we explore more about its performance in different numbers of training data scenarios.   


\begin{figure*}[] \label{figure3}
\begin{center}
\includegraphics[width=1\linewidth]{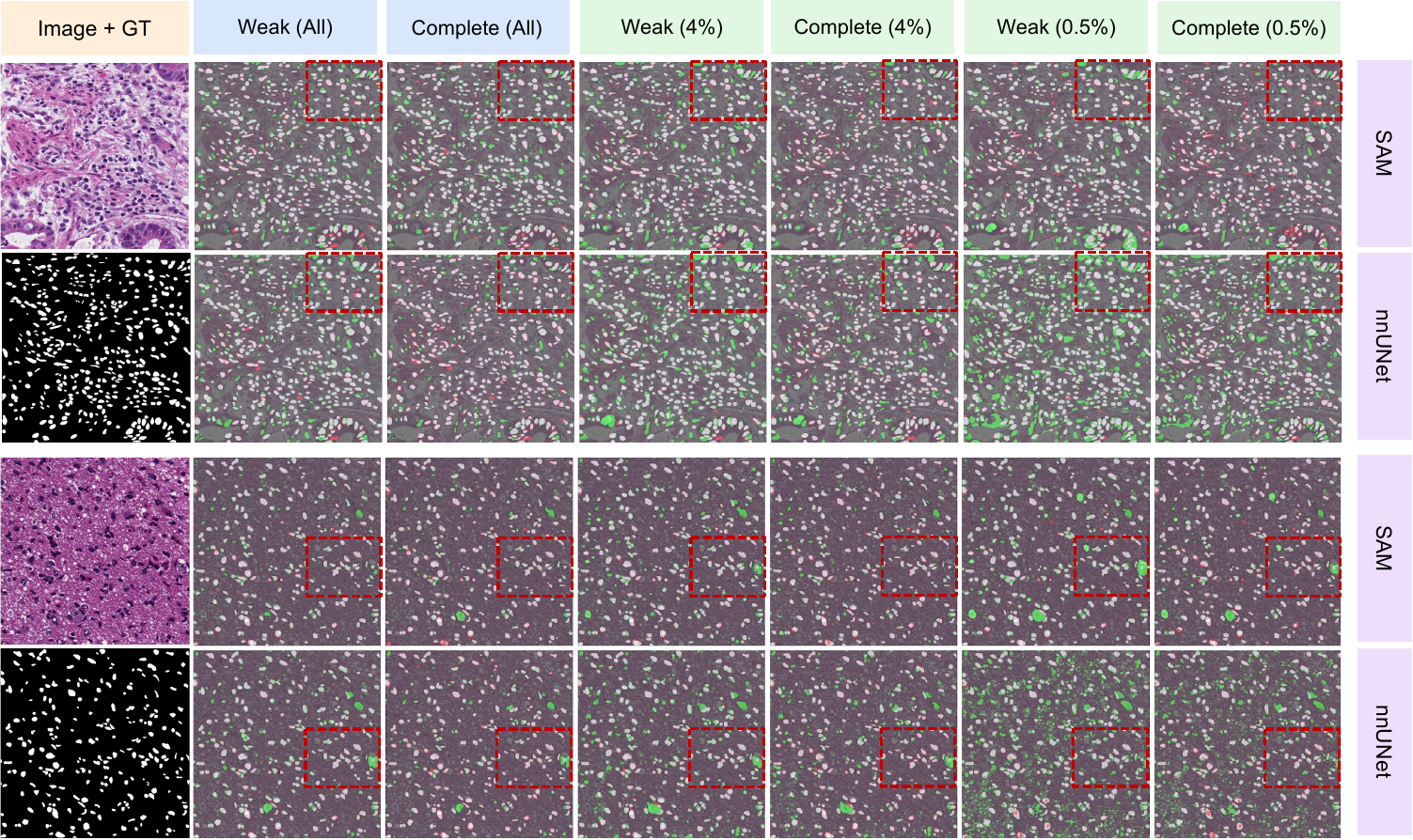}
\end{center}
   \caption{Qualitative results of nuclei segmentation of different models trained by different numbers of annotated data. Upper row: proposed method; lower row: nnUNet. The mask region in green color indicates the false positive of prediction, the white color indicates the true positive, and the red color indicates the false negative.}
\label{fig2:Cross-modal}
\end{figure*}

\section{DATA AND EXPERIMENTS}

\subsection{Data and Task}
In this work, we employ the MICCAI 2018 Monuseg dataset~\cite{kumar2019multi}. It consists of 30 training images and 14 testing images, all with dimensions of 1000$\times$1000 pixels. Each image is accompanied by corresponding masks of nuclei. To ensure a fair and comparable evaluation, we use the same data split as a recent study~\cite{li2022lvit}. The 30 training images are divided into two subsets, with 24 images assigned to the training set and the remaining 6 images forming the validation set. To evaluate the model performance of nucleus segmentation, Dice, AUC, Recall, Precision, best F1 (maximized F1 score at the optimal threshold), IoU (Intersection over Union) and ADJ (Adjusted Rand Index) are calculated.

\begin{table*}[] \label{table1}
\caption{Comparison with other SOTA methods when using different numbers of training data with weak or complete annotation. The best performance is highlighted in red color, while the second-best performance is highlighted in blue color.}
\centering
\begin{tabular}{|
>{\columncolor[HTML]{FFFFFF}}l |
>{\columncolor[HTML]{FFFFFF}}l |
>{\columncolor[HTML]{FFFFFF}}l |
>{\columncolor[HTML]{FFFFFF}}r 
>{\columncolor[HTML]{FFFFFF}}r 
>{\columncolor[HTML]{FFFFFF}}r |}
\hline
Label & Method & Training Data & \multicolumn{1}{l|}{\cellcolor[HTML]{FFFFFF}Dice} & \multicolumn{1}{l|}{\cellcolor[HTML]{FFFFFF}IoU} & \multicolumn{1}{l|}{\cellcolor[HTML]{FFFFFF}ADJ} \\ \hline
\cellcolor[HTML]{FFFFFF} & Xie et al \cite{xie2020instance} & All & \multicolumn{1}{l}{\cellcolor[HTML]{FFFFFF}-} & \multicolumn{1}{l}{\cellcolor[HTML]{FFFFFF}-} & {\color[HTML]{980000} \textbf{70.63\%}} \\ \cline{3-6} 
\cellcolor[HTML]{FFFFFF} &  & 30\% & \multicolumn{1}{l}{\cellcolor[HTML]{FFFFFF}} & \multicolumn{1}{l}{\cellcolor[HTML]{FFFFFF}} & 60.31\% \\ \cline{3-6} 
\cellcolor[HTML]{FFFFFF} &  & 10\% & \multicolumn{1}{l}{\cellcolor[HTML]{FFFFFF}-} & \multicolumn{1}{l}{\cellcolor[HTML]{FFFFFF}-} & 55.01\% \\ \cline{2-6} 
\cellcolor[HTML]{FFFFFF} & LViT \cite{li2022lvit} & All & 80.33\% & 67.24\% & \multicolumn{1}{l|}{\cellcolor[HTML]{FFFFFF}-} \\ \cline{3-6} 
\cellcolor[HTML]{FFFFFF} &  & 25\% & 79.94\% & 66.80\% & \multicolumn{1}{l|}{\cellcolor[HTML]{FFFFFF}-} \\ \cline{2-6} 
\cellcolor[HTML]{FFFFFF} & BEDs \cite{li2021beds} & All+More & 81.77\% & \multicolumn{1}{l}{\cellcolor[HTML]{FFFFFF}-} & \multicolumn{1}{l|}{\cellcolor[HTML]{FFFFFF}-} \\ \cline{2-6} 
\cellcolor[HTML]{FFFFFF} & nnUNet \cite{isensee2021nnu} & All & 82.44\% & {\color[HTML]{980000} \textbf{69.76\%}} & 70.28\% \\ \cline{3-6} 
\cellcolor[HTML]{FFFFFF} &  & 4\% & 79.20\% & 65.40\% & 65.80\% \\ \cline{2-6} 
\cellcolor[HTML]{FFFFFF} & Proposed & All & {\color[HTML]{980000} \textbf{82.54\%}} & {\color[HTML]{0000FF} \textbf{69.74\%}} & {\color[HTML]{0000FF} \textbf{70.36\%}} \\ \cline{3-6} 
\multirow{-10}{*}{\cellcolor[HTML]{FFFFFF}Complete} &  & 4\% & 81.34\% & 68.10\% & 68.67\% \\ \hline
\cellcolor[HTML]{FFFFFF} & nnUNet  \cite{isensee2021nnu}  & All & 82.12\% & 69.35\% & 69.74\% \\ \cline{3-6} 
\cellcolor[HTML]{FFFFFF} &  & 4\% & 79.13\% & 65.27\% & 65.62\% \\ \cline{2-6} 
\cellcolor[HTML]{FFFFFF} & Proposed & All & {\color[HTML]{0000FF} \textbf{82.46\%}} & 69.73\% & 70.24\% \\ \cline{3-6} 
\multirow{-4}{*}{\cellcolor[HTML]{FFFFFF}Weak} &  & 4\% & 80.99\% & 67.60\% & 68.14\% \\ \hline
\end{tabular}
\end{table*}


\begin{table*}[] \label{table2}

\caption{Comparison of different methods trained by different numbers of weakly/completely annotated data}

\centering
\begin{tabular}{|
>{\columncolor[HTML]{FFFFFF}}l |
>{\columncolor[HTML]{FFFFFF}}l |
>{\columncolor[HTML]{FFFFFF}}l |
>{\columncolor[HTML]{FFFFFF}}r 
>{\columncolor[HTML]{FFFFFF}}r 
>{\columncolor[HTML]{FFFFFF}}r 
>{\columncolor[HTML]{FFFFFF}}r 
>{\columncolor[HTML]{FFFFFF}}r 
>{\columncolor[HTML]{FFFFFF}}r 
>{\columncolor[HTML]{FFFFFF}}r |}
\hline
Label & Method & Training Data & \multicolumn{1}{l|}{\cellcolor[HTML]{FFFFFF}Dice} & \multicolumn{1}{l|}{\cellcolor[HTML]{FFFFFF}AUC} & \multicolumn{1}{l|}{\cellcolor[HTML]{FFFFFF}Recall} & \multicolumn{1}{l|}{\cellcolor[HTML]{FFFFFF}Precision} & \multicolumn{1}{l|}{\cellcolor[HTML]{FFFFFF}bestF1} & \multicolumn{1}{l|}{\cellcolor[HTML]{FFFFFF}IoU} & \multicolumn{1}{l|}{\cellcolor[HTML]{FFFFFF}ADJ} \\ \hline
\cellcolor[HTML]{FFFFFF} & \cellcolor[HTML]{FFFFFF} & All & 82.44\% & 96.04\% & 82.82\% & 82.55\% & 83.21\% & 69.76\% & 70.28\% \\ \cline{3-3}
\cellcolor[HTML]{FFFFFF} & \cellcolor[HTML]{FFFFFF} & 4\% & 79.20\% & 94.10\% & 84.90\% & 74.60\% & 79.70\% & 65.40\% & 65.80\% \\ \cline{3-3}
\cellcolor[HTML]{FFFFFF} & \multirow{-3}{*}{\cellcolor[HTML]{FFFFFF}nnUNet} & 0.50\% & 76.23\% & 92.49\% & 86.79\% & 68.61\% & 77.97\% & 61.35\% & 61.86\% \\ \cline{2-10} 
\cellcolor[HTML]{FFFFFF} & \cellcolor[HTML]{FFFFFF} & All & 82.54\% & 97.17\% & 84.74\% & 80.81\% & 83.04\% & 69.74\% & 70.36\% \\ \cline{3-3}
\cellcolor[HTML]{FFFFFF} & \cellcolor[HTML]{FFFFFF} & 4\% & 81.34\% & 95.50\% & 84.92\% & 78.53\% & 81.90\% & 68.10\% & 68.67\% \\ \cline{3-3}
\multirow{-6}{*}{\cellcolor[HTML]{FFFFFF}Complete} & \multirow{-3}{*}{\cellcolor[HTML]{FFFFFF}Proposed} & 0.50\% & 78.16\% & 94.40\% & 84.57\% & 73.51\% & 79.17\% & 63.79\% & 64.30\% \\ \hline
\cellcolor[HTML]{FFFFFF} & \cellcolor[HTML]{FFFFFF} & All & 82.12\% & 95.65\% & 91.26\% & 74.98\% & 83.68\% & 69.35\% & 69.74\% \\ \cline{3-3}
\cellcolor[HTML]{FFFFFF} & \cellcolor[HTML]{FFFFFF} & 4\% & 79.13\% & 94.74\% & 92.76\% & 69.22\% & 81.54\% & 65.27\% & 65.62\% \\ \cline{3-3}
\cellcolor[HTML]{FFFFFF} & \multirow{-3}{*}{\cellcolor[HTML]{FFFFFF}nnUNet} & 0.50\% & 75.00\% & 92.64\% & 93.36\% & 61.99\% & 77.57\% & 58.81\% & 59.18\% \\ \cline{2-10} 
\cellcolor[HTML]{FFFFFF} & \cellcolor[HTML]{FFFFFF} & All & 82.46\% & 97.32\% & 89.47\% & 76.78\% & 83.39\% & 69.73\% & 70.24\% \\ \cline{3-3}
\cellcolor[HTML]{FFFFFF} & \cellcolor[HTML]{FFFFFF} & 4\% & 80.99\% & 95.22\% & 88.45\% & 75.09\% & 81.87\% & 67.60\% & 68.14\% \\ \cline{3-3}
\multirow{-6}{*}{\cellcolor[HTML]{FFFFFF}Weak} & \multirow{-3}{*}{\cellcolor[HTML]{FFFFFF}Proposed} & 0.50\% & 78.73\% & 94.30\% & 87.04\% & 72.48\% & 79.82\% & 64.57\% & 65.02\% \\ \hline
\end{tabular}
\end{table*}

\subsection{Experiment Setting}
In this work, we designed 3 sets of experiments to explore the performance of finetuned SAM on the nucleus segmentation task. 

\textbf{1) Finetuned by complete annotation or weak annotation.} For complete annotation, the pixel-wise complete annotations were provided for training data to finetune the pretrained SAM model. As for the weak annotation, only the bounding boxes of nuclei were provided. In this work, the bounding boxes were automatically prepared by using the complete masks. And then, these bounding boxes were used as the prompts in the pretrained SAM to generate pixel-level pseudo labels for finetuning. 

\textbf{2) Finetuned by different numbers of annotated data.}
To evaluate the performance of the proposed pipeline finetuned with different numbers of data, we adjusted the number of annotated images and the area of annotated regions. The complete training set contains 24 1000$\times$1000 image patches with corresponding annotations. In the 4\%  training data set, only a 200$\times$200 random patch was selected from each large patch for annotation. To keep the parameters unchanged for a fair comparison, the rest area without annotation was set to intensity 0. In the extreme cases, only 3 patches (1 from each image) in the size of 200$\times$200, taking up 0.5\% of the original complete dataset, were randomly selected. 

\textbf{3) Comparison with other SOTA.}
In this study, we conducted a performance comparison between our proposed pipeline and other SOTA methods. LViT \cite{li2022lvit} is a recently proposed model integrating language information and images for annotation and achieved SOTA performance in Monuseg dataset. We followed their data splits in our experiments. BEDs~\cite{li2021beds} 
integrated the self-ensemble and testing
stage stain augmentation mechanism in  UNet models for nuclei segmentation. Although more data were used for training, the model was evaluated in the same testing set. So, the comparable performance results of LViT~\cite{li2022lvit} and BEDs~\cite{li2021beds} are from their paper. For the task of learning from a small annotated dataset, Xie et al.~\cite{xie2020instance} proposed to use self-supervised learning to utilize the unlabeled data and achieved better results when only a small number of annotated training data were available. Their proposed method was implemented on the same Monuseg training and testing dataset, so their results were reported here for comparison with ours. In addition, nnUNet, as a popular benchmark in the medical image segment, was run by ourselves on the Monuseg with the same dataset setting as our proposed pipeline. To ensure a fair comparison, we used the default settings of nnUNet and trained it for the default 1000 epochs.


About other settings in our proposed pipeline, the ViT-H backbone was used in both the annotation generation and fine-tuning stages. The training would early stop if the validation loss did not decrease for consecutive 40 epochs. Without specific mention, other default parameters and settings in SAM-adapter\cite{chen2023sam} were kept. All experiments were repeated three times for average evaluation values. An RTX A6000 was used to run these experiments.


\section{Results and Discussion}
Table~\textcolor{blue}{1} shows the comparison of our results with other SOTA methods. The best performance was observed when training with the whole training set and complete annotation. Notably, the nnUNet~\cite{isensee2021nnu}, our proposed method, and Xie's method~\cite{xie2020instance} demonstrated similar performance in this scenario. However, when training with a reduced number of annotated data, our proposed method exhibited a smaller drop in performance compared to other methods and achieved superior results. Additionally, when employing weak labels for training, the proposed method consistently outperformed other methods, maintaining the highest performance. Table~\textcolor{blue}{2} and Fig.~\ref{fig2:Cross-modal} provide a comprehensive view of the evaluation metrics and show the performance under extreme cases where only 0.5\% of the training set data is available. In various evaluation metrics such as Dice, AUC, Precision, bestF1, IoU, and ADJ, the proposed method consistently outperformed nnUNet across different settings. This was particularly evident when utilizing limited and poorly annotated data. However, when training with fewer and weakly annotated data, the proposed method exhibited a lower Recall value compared to nnUNet. Despite this, nnUNet displayed a more aggressive approach to segmenting nuclei, resulting in significantly lower precision and other metrics.

\section{Conclusion}
In summary, we introduce an efficient and effective pipeline that leverages a pretrained self-attention mechanism (SAM) for nuclei segmentation with limited annotation. The experiments show the capability of the pretrained SAM model to generate pseudo labels from weak annotations and subsequently finetune with these pseudo labels during the inference phase. This approach achieves competitive performance when compared to SOTA methods while significantly reducing the burden of manual annotation. This pipeline holds great significance in real-world applications of nuclei segmentation, as it offers a practical solution that minimizes annotation efforts without compromising on segmentation accuracy.

\section{Acknowledgements}
This work is supported by the Leona M. and Harry B. Helmsley Charitable Trust grant G-1903-03793, NSF CAREER 1452485.




\bibliography{reference} 
\bibliographystyle{IEEEtran} 
\end{document}